# UF-HOBI at "Discharge Me!": A Hybrid Solution for Discharge Summary Generation Through Prompt-based Tuning of GatorTronGPT Models

Mengxian Lyu[1, *], Cheng Peng[1, *], Daniel Paredes[1], Ziyi Chen[1],
Aokun Chen[1], Jiang Bian[1, 2], Yonghui Wu[1, 2]

[1]Department of Health Outcomes and Biomedical Informatics, University of Florida
[2]Cancer Informatics Shared Resource, University of Florida Health Cancer Center
{lvmengxian, c.peng, dparedespardo, chenziyi, chenaokun1990, bianjiang, yonghui.wu}@ufl.edu

## Abstract

Automatic generation of discharge summaries presents significant challenges due to the length of clinical documentation, the dispersed nature of patient information, and the diverse terminology used in healthcare. This paper presents a hybrid solution for generating discharge summary sections as part of our participation in the "Discharge Me!" Challenge at the BioNLP 2024 Shared Task. We developed a two-stage generation method using both extractive and abstractive techniques, in which we first apply name entity recognition (NER) to extract key clinical concepts, which are then used as input for a prompt-tuning-based GatorTronGPT model to generate coherent text for two important sections including "Brief Hospital Course" and "Discharge Instructions". Our system was ranked 5th in this challenge, achieving an overall score of 0.284. The results demonstrate the effectiveness of our hybrid solution in improving the quality of automated discharge section generation.

## 1 Introduction

The discharge summary is one of the most crucial documents that capture patients' present illness, diagnostic findings, therapeutic procedures, and follow-up instructions(Lenert et al., 2014). Timely, high-quality discharge records can remarkably reduce the risk of patient readmissions, ensuring continuous and coordinated patient care, supporting the decision-making process, and bridging the information gap between healthcare providers (Kripalani et al., 2007; Li et al., 2013; van Walraven et al., 2002). However, manually writing discharge summaries is time-consuming and error-prone, given the complexity of clinical information, the dispersed nature of patient details, and the increasing burden of clinical documentation (S. Lin et al., 2010).

Despite the recent success of large language models (LLMs) in natural language process (NLP) (Karabacak & Margetis, 2023), it's still challenging for LLMs to summarize critical patient information from a long clinical document, which often exceeds the maximum input length of LLMs, making it challenging for LLMs to process all relevant information at once. This leads to truncated input and potentially low-quality content. Additionally, excessive tokens can overwhelm LLM's capacity to focus on important patient information, affecting both the quality and coherence of the generated summaries (Van Veen et al., 2023).

To counter these challenges, we propose a two-step approach to generate the target sections. The process begins with a rule-based segmentation of original discharge summaries into individual sections. We manually reviewed a subset of notes in the training set to identify the sections that contain important information related to the two target sections. Next, we apply the GatorTron (Yang et al., 2022) model, fine-tuned on the 2010 i2b2 Challenge (Uzuner et al., 2011) dataset to extract critical clinical concepts related to problems, treatments, and lab tests in selected sections. The extracted concepts are then concatenated with selected sections, serving as input for GatorTronGPT to generate "Brief Hospital Course" and "Discharge Instructions" sections using soft prompt-tuned GatorTronGPT (Peng et al., 2023). Compared with directly using the original long document, our hybrid approach remarkably reduces the number of input tokens and helps LLMs focus on critical patient information to generate good-quality summaries.

---

[*] Equal contribution

## 2   Related Work

Automatic Text Summarization (ATS) is a critical Natural Language Processing (NLP) task that focuses on generating concise summaries from a long document. By extracting or abstracting essential information, ATS provides comprehensive yet significantly shorter versions of source content. There are two primary approaches for ATS, including extractive - which identifies and selects essential sentences directly from the text, and abstractive - which generates new content that conveys the original meaning(Sharma & Sharma, 2022). Both techniques play an important role in effectively condensing information, making it easier to digest while retaining the core message.

The advance of transformer-based large language models (LLMs) has revolutionized ATS. Through pre-training on extensive amounts of text, LLMs demonstrate good ability in transfer learning, few-shot learning, and zero-shot learning and achieve state-of-the-art performance in both extractive and abstractive summarization. Language models like BERT (Devlin et al., 2019) and GPT-3 (Brown et al., 2020) have been widely used in understanding and generating text. BERT's bidirectional architecture is adept at contextual comprehension, which is useful for extracting original text from context. GPT-3, an autoregressive transformer, is better at generating abstract contents that are coherent and contextually relevant to the original text. However, clinical summarization is still challenging due to the complex, specialized vocabulary and long text documents, which hamper the performance of ATS due to token limitations and dense information(Karabacak & Margetis, 2023). To address these challenges, hybrid methods that integrate both extractive and abstractive techniques are increasingly being used. (Krishna et al., 2021) proposed a method leveraging the extractive summarization model's distill ability to extract essential information from long documents and an abstractive summarization pipeline to generate concise Subjective, Objective, Assessment, and Plan (SOAP) notes.

Prompt-based learning is another technology that improved text generation by providing LLMs with instructional cues embedded in the input data. "Hard prompts" (or discrete prompts) and "soft prompts" (or continuous prompts) are two types of prompts used in prompt-based methods. Due to the labor-intensive nature and potential for miscommunication between humans and models, hard prompts often struggle to achieve optimal performance in guiding model behavior(Lester et al., 2021). In contrast, soft prompts, which are embeddings that can be optimized during training, have a better ability to instruct LLMs for ATS. Recent studies have shown that prompt-tuning can effectively instruct LLMs for various NLP tasks. P-tuning, a specific form of prompt tuning, further optimizes trainable continuous vectors to capture task-specific knowledge without updating model weights (X. Liu et al., 2023).

Retrieval augmented generation (RAG) has been rapidly developing in recent years as a key technology in advancing LLMs by retrieving relevant documents through semantic similarity calculation (Lewis et al., 2020). Recent studies have shown the effectiveness of RAG for summarization of computer codes in the general domain (S. Liu et al., 2020; Parvez et al., 2021). RAG-based summarization uses a "Retriever" to first identify the sentences that meet the summarization instructions through semantic similarity calculation, which will be used as the input for a "Generator" to generate a shorter summary. Thus, the "Retriever" and "Generator" are the key components.

## 3   Dataset

The "Discharge Me!" challenge dataset (Xu et al., 2024)

[1] is curated from the MIMIC-IV database (Johnson et al., 2023) and features over 109,000 ED visits. Each record includes ICD-9 or ICD-10 diagnosis codes, chief complaints, at least one radiology report, and a discharge summary with "Brief Hospital Course" and "Discharge Instructions". The dataset was split into training (68,785 samples), validation (14,719 samples), phase I testing (14,702 samples), and phase II testing (10,962 samples) subsets. The phase II testing dataset will serve as the final test set. All datasets and tables are derived from the MIMIC-IV submodules.

The challenge focuses on the automated generation of the "Brief Hospital Course" and "Discharge Instructions" sections. Table 1 shows the items from different sources. All sources of data in the training and validation sets are allowed to use for model training except the two target sections.

| Item | Total Count |
| --- | --- |
| Visits | 109,168 |
| Discharge Summaries | 109,168 |
| Radiology Reports | 409,359 |
| ED Stays | 109,403 |
| ED Diagnoses | 218,376 |

Table 1: Source of Dataset items

## 4 Methods

Triggered by the recent RAG-based summarization methods, we developed a hybrid solution that is composed of a "Retriever" and a "Generator". We fine-tuned an encoder-only clinical LLM, GatorTron, as the retriever to identify important clinical concepts, which were used by the Generator, GatorTronGPT, to generate the target sections. To reduce the length of the input, we used a rule-based method to segment the notes into individual sections. We manually examined several notes from the training set to identify (1) a subset of sections that are directly related to the target sections, and (2) a subset of sections useful but not directly relevant to the target sections. The input was reconstructed by concatenating: (1) original text from the sections directly related to the target sections, (2) Clinical concepts extracted using GatorTron from the sections useful but not directly related to the target sections, and (3) diagnosis descriptions. To instruct GatorTronGPT to generate target sections, we explored four strategies by combining different tuning methods and input construction methods: (1) traditional fine-tuning using original inputs, (2) fine-tuning using our reconstructed inputs, (3) prompt-based tuning (p-tuning) using original inputs, and (4) p-tuning using our reconstructed inputs.

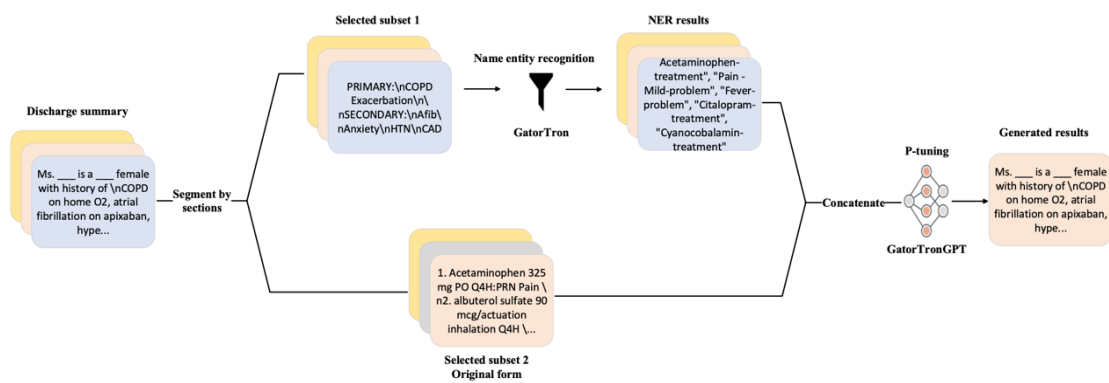

Figure 1: Overview of our summary generation pipeline

The following sections highlight the approach that demonstrated the best performance—p-tuning using both NER results and original texts. This method integrates several advanced techniques and models to optimize outcomes. As illustrated in Figure 1, our best strategy combines the generative capabilities of state-of-the-art clinical large language models, the extractive ability of NER systems, and efficient instruction using soft-prompt tuning techniques. The experimental results show

---
[1]https://physionet.org/content/discharge-me/1.3/

that our approach can generate coherent contexts for the two important clinical sections.

The following subsections describe the models and methods used in our study, including the model architectures, training strategies, and evaluation metrics.

### 4.1 Large Language Models

**GatorTron** (Yang et al., 2022), a BERT-style large clinical language model, pretrained on over 90 billion words. This extensive corpus included more than 80 billion words from 290 million clinical notes sourced from the University of Florida (UF) Health System, encompassing patient records from 2011 to 2021 across more than 126 clinical departments and approximately 50 million encounters. These notes spanned various healthcare settings, such as inpatient, outpatient, and emergency department visits.

**GatorTronGPT** (Peng et al., 2023) is a generative clinical large language model specifically developed for medical research and healthcare applications. It was trained on 277 billion words, including 82 billion words of de-identified clinical text from the University of Florida (UF) Health and 195 billion words of general English text. Utilizing the GPT-3 architecture with up to 20 billion parameters, GatorTronGPT demonstrated superior performance in biomedical natural language processing tasks such as relation extraction and question answering. Prior studies have demonstrated GatorTronGPT's capability to generate precise and contextually pertinent summaries from doctor-patient dialogues. (Lyu et al., 2024). In this study, we deploy both the GatorTronGPT-5B and GatorTronGPT-20B models to further explore their efficacy in addressing abstractive summarization tasks.

### 4.2 Named Entity Recognition

In our approach, we applied Named Entity Recognition (NER) before the abstractive summarization step to focus LLMs on critical clinical concepts in the notes. We employed GatorTron, fine-tuned on the 2010 i2b2 dataset, including annotated concepts for problems, treatments, and lab tests, to capture important patient information from clinical notes. This enhanced the focus of GatorTronGPT on using important healthcare information to generate note sections.

### 4.3 P-tuning for Abstractive Summarization

To enhance the performance of GatorTronGPT for abstractive summarization, we adopted p-tuning methods. Specifically, we incorporated "soft prompts" as trainable variables to instruct GatorTronGPT. During the tuning process, the GatorTronGPT weights remain unchanged, and only the parameters of the soft prompts are updated. This technique involves adding a sequence of virtual tokens to the input, which are represented by trainable embeddings dynamically adjusted through Multi-Layer Perceptron (MLP) and Long Short-Term Memory (LSTM) networks.

The P-tuning method allows the model to utilize its extensive pre-trained weights while fine-tuning it to a specific task of generating precise and contextually relevant summaries from input texts. Since only the parameters of the soft prompts were updated in backpropagation and the parameters of GatorTronGPT were not updated, our solution provides a cost-efficient solution to instruct LLMs for ATS.

In this study, we implemented P-tuning using both GatorTronGPT-5B and GatorTronGPT-20B models. The training objective is to minimize the cross-entropy loss, calculated based on the discrepancy between the model-generated summaries and the gold-standard summaries. This objective ensures that the generated summaries are both precise and contextually relevant.

### 4.4 Automatic Evaluation

We used the official evaluation metrics released by the challenge organizers to evaluate our generated sections. Based on the textual similarity and factual correctness, including BLEU-4 (Papineni et al., 2002), Rouge (C.-Y. Lin, 2004), BERTScore (Zhang et al., 2019), METEOR (Lavie., n.d.), AlignScore (Zha et al., 2023), and MEDCON (Yim et al., 2023), the final results are scored separately for each target section ("Brief Hospital Course" and "Discharge Instructions"), and the mean score for each metric is calculated across all test samples. The mean of the scores for each metric across both target sections is then computed, and the overall system score is the mean of these metric means.

### 4.5 Human Evaluation

Three clinicians evaluated a subset of 25 samples from the test phase. The evaluations using five-point Likert scale measurements focus on

Completeness, Correctness, Readability, and Holistic Comparison to the Reference Text. Scores from the three clinicians were averaged for each sample and then averaged across the 25 samples. This yielded seven total scores: four for the Brief Hospital Course (completeness, correctness, readability, and overall) and three for Discharge Instructions (completeness, correctness, and overall).

## 5 Experiments

### 5.1 Data Exploration

We manually reviewed a subset of notes from different sources. All the discharge summaries that contain a "Brief Hospital Course" and a "Discharge Instructions" section were used. Each visit is defined by a unique "hadm_id" and is associated with a corresponding discharge summary with at least one radiology report.

We performed a statistical analysis to determine the average length and discovered that nearly 15% percent of discharge summaries exceed the maximum input length of our generative model. This finding underscores the need for effective truncation or summarization strategies to ensure compatibility with the model's input constraints.

### 5.2 Data Preprocessing

We performed the following steps to facilitate the model training for generating discharge summary sections.

#### 5.2.1 Data Segmentation

| Segmented Sections |
| --- |
| Chief Complaint |
| Major Surgical or Invasive Procedure |
| History of Present Illness |
| Past Medical History |
| Social History |
| Family History |
| Physical Exam |
| Pertinent Results |
| Brief Hospital Course |
| Medications on Admission |
| Discharge Medications |
| Discharge Disposition |
| Discharge Diagnosis |
| Discharge Condition |
| Discharge Instructions |

Table 2: Segmentation results for discharge notes sections

Segmentation is important to isolate specific narrative blocks relevant to different aspects of target sections. We applied a rule-based method to segment the discharge summaries into clinical sections, leveraging the existing structure of clinical notes. Specifically, we manually created a list of notes section names to split each section. These sections included "Chief Complaint", "Major Surgical or Invasive Procedure", "History of Present Illness", and several others leading to discharge summary sections. Table 2 shows the data segmentation result for the discharge summary.

#### 5.2.2 Data Selection

The "Brief Hospital Course" section is used to synthesize a detailed narrative of the patient's hospital stay, emphasizing the sequence of medical events, interventions, and outcomes. The "Discharge Instructions" section is used to produce clear and concise guidelines for post-hospital care, ensuring that instructions are patient-centric, easy to understand, and aligned with best-practice recovery protocols(Searle et al., 2023).

We manually identified the note sections that may contain the required information for the two target sections. For each target section, we reviewed randomly sampled notes from the training set to identify relevant note sections that contain important information related to the target section. We divided the selected sections into two subsets: subset 1, which were processed by GatorTron to extract important clinical concepts, and subset 2, which were directly used in the input as they are directly related to the target sections. Table 3 in the Appendix provides the sections we selected for generating the two target sections.

#### 5.2.3 Identify Critical Patient Information using GatorTron

We fine-tuned the GatorTron model using the i2b2 2010 challenge dataset following the default training and test settings. We applied fine-tuned GatorTron to recognize the following clinical concepts:

- PROBLEM: including clinical conditions, symptoms, and diagnoses, which identify the patient's primary and secondary health issues
- TREATMENT: including procedures, medications, and other therapeutic interventions, which detailing the medical and surgical management of the patient.

- TEST: including diagnostic tests and their results, which are essential for the diagnosis, monitoring, and management of health conditions.

We processed the separated sections individually to generate a set of clinical concepts for different sections.

### 5.3 P-tuning for Note Section Generation

**Prompt construction** We constructed a general instruction template for fine-tuning the GatorTronGPT models. The prompts templet is structured as follows: "<|VIRTUAL_PROMPT|> Input: {input}\n Output:{output}", where placeholder "<|VIRTUAL_PROMPT|>" represent soft prompt which was randomly initialized at the beginning and updated during the p-tuning.

To ensure the generation quality, we carefully designed an input prompt to focus GatorTronGPT. Each input prompt begins with clear instructions to guide the model: "*Given the following concepts and text extracted from each section in a discharge summary, generate the section 'Discharge Instructions'*". We used this instruction to instruct GatorTronGPT to generate the target sections properly.

To integrate the selected sections as input, we extracted all the clinical concepts using GatorTron from the selected subset 1 and concatenated them with commas. Different sections are isolated using "\n". The output is the target section specified in the input instruction, ensuring the model focuses on generating the appropriate discharge section. Table 8 in the Appendix shows the input prompt we construct for different target sections.

**Experimental Setting** We adopted a grid search to optimize the hyperparameters, including the training hyperparameter learning rate, the training batch size and the P-tuning virtual token length, and the inference hyperparameter temperature and the value of top p in nucleus sampling. We used the training set provided in this challenge to p-tuning GatorTronGPT. The best models were selected according to the cross-validation performances measured by the overall score based on the evaluation metrics provided by the challenge. We used the following parameters in our best performance: a global batch size of 64, a learning rate of 0.0001, a virtual token length of 50, a temperature of 0.2, and a top p of 0.6. All experiments were conducted using 8 Nvidia A100-80G GPUs.

## 6 Results

### 6.1 Extract Clinical Concepts Using GatorTron

We fine-tuned GatorTron using the 2010 i2b2 datasets to extract problems, treatments, and lab tests from the selected sections to reduce the input length. Table 3 compares the average input length between the original contents, and GatorTron extracted concepts and compares them with the original content. Using GatorTron to extract concepts reduced the input length by 80% on average.

| Target section | Split | Original Content | NER result |
|---|---|---|---|
| Brief Hospital Course | Trian | 2921 | 450 |
| | Valid | 2925 | 446 |
| | Test | 2910 | 445 |
| Discharge Instructions | Trian | 2921 | 401 |
| | Valid | 2025 | 405 |
| | Test | 2910 | 400 |

Table 3: Average length among data splits

| Original Content | NER result |
|---|---|
| - prior paramedian pontine infarct (___) \n- right-sided lenticulostriate territory infarct ___ \n- Hypertension as per prior medical records(patient denies)\n- Dyslipidemia \n- Colon cancer 2/p right colectomy in ___ with prolonged\nstuttering course of adjuvant chemotherapy (diagnosed in setting\nof GI bleeding)\n- Cholecystectomy for chronic cholecystitis and gallstones in\n___ \n- Diverticulosis\n- Hemorrhoids | prior paramedian pontine infarct, right-sided lenticulostriate territory infarct, Hypertension, Dyslipidemia, Colon cancer, right colectomy, adjuvant chemotherapy, GI bleeding, Cholecystectomy, chronic cholecystitis, gallstones, Diverticulosis, Hemorrhoids |

Table 4: An example of original content and GatorTron extracted concepts.

Table 4 provides an example of the original text and the GatorTron-extracted concepts. GatorTron can extract critical concepts with important clinical meaning to facilitate section generation using GatorTronGPT models.

### 6.2 Target Note Section Generation

Table 5 compares different strategies to generate the target note sections. The GatorTronGPT-20B model consistently outperformed the GatorTronGPT-5B across all evaluation metrics, achieving the highest overall score of 0.2885. Furthermore, p-tuning demonstrated better performance than traditional fine-tuning methods. Notably, our strategy to combine NER results with the original text consistently achieved higher scores across all evaluation metrics for both models under different training strategies. Both GatorTronGPT-5B and GatorTronGPT-20B showed remarkable improvements with p-tuning when using NER results combined with the original text. GatorTronGPT-20B achieved the best BLEU score of 0.1211 and BERTScore of 0.3894.

| Model | Strategy | Input | BLEU-4 | Rouge-1 | Rouge-2 | Rouge-L | BERTScore | Meteor | Align | Medcon | Overall |
|---|---|---|---|---|---|---|---|---|---|---|---|
| GatorTronGPT-5B | FT | All Text | 0.0374 | 0.1628 | 0.0431 | 0.1420 | 0.2931 | 0.2440 | 0.1688 | 0.2796 | 0.1714 |
| | | NER + Text | 0.0739 | 0.3181 | 0.1252 | 0.2118 | 0.3330 | 0.2518 | 0.2440 | 0.3166 | 0.2342 |
| | PT | All Text | 0.0538 | 0.1941 | 0.0391 | 0.1397 | 0.3099 | 0.2872 | 0.1890 | 0.2849 | 0.1872 |
| | | NER + Text | 0.0817 | 0.3570 | 0.1076 | 0.2352 | 0.3679 | 0.3213 | 0.2567 | 0.3193 | 0.2558 |
| GatorTronGPT-20B | FT | All Text | 0.0498 | 0.1829 | 0.0643 | 0.1689 | 0.2997 | 0.2459 | 0.1738 | 0.2987 | 0.1855 |
| | | NER + Text | 0.0956 | 0.3402 | 0.1715 | 0.2496 | 0.3794 | 0.2446 | 0.2498 | 0.3440 | 0.2593 |
| | PT | All Text | 0.0483 | 0.2671 | 0.1059 | 0.1798 | 0.2791 | 0.1908 | 0.2636 | 0.3325 | 0.2084 |
| | | NER + Text | **0.1211** | **0.3958** | **0.1790** | **0.2699** | **0.3894** | **0.2985** | **0.2835** | **0.3705** | **0.2885** |

Table 5: Results of GatorTronGPT using different training strategies (FT: fine-tuning, PT: p-tuning).

### 6.3 Generation Result Analysis

Table 9 in the Appendix provides examples generated by GatorTronGPT and compared with the corresponding ground truth.

For the "Brief Hospital Course" section, the section generated by GatorTronGPT captures the patient's conditions and history accurately but rewrites them in a list format different from the narrative format in the original text. The list format is concise and clear but with a lower readability compared to the narrative format of the ground truth.

For the "Brief Hospital Course" section, the text generated by GatorTronGPT accurately captures the patient's conditions and history, showcasing the model's ability to understand and summarize complex medical information. The list format used in the generated text is concise and clear, making it easy to identify key information quickly. While the ground truth uses a narrative format that offers a more cohesive flow, the list format enhances the document's usability in a clinical setting by high-lighting essential details.

For the "Discharge Instructions" section, the section generated by GatorTronGPT captured most of the key information from the ground truth and covered the admission reasons (nausea, heart fail-ure), treatment (diuretics), and high blood pressure medications. The generated result simplifies some details (e.g., "heart failure exacerbation" instead of "too much fluid in your body (heart failure)"). This simplification demonstrates a strength in producing clear and concise instructions.

### 6.4 Human Evaluation

The organizer picked up 25 samples from the submitted results and recruited three clinicians for manual evaluation. For each sample, the three clinicians evaluated the Readability, Correctness, and Completeness using scores from 1 to 5, where 1 indicates the worst score and 5 the best score. The overall score was derived by calculating the average score. Table 6 shows the human evaluation scores. For the Brief Hospital Course, we achieved a Correctness score of 3.3600, indicating that our content contained few inaccuracies and was unlikely to impact future care adversely. Additionally, the Readability score of 2.7067 shows that our text, while slightly harder to read than the reference, maintained a reasonable level of clarity. For the Discharge Instructions, our Completeness score of 3.0133 highlights our ability to capture a significant portion of important information, and a Correctness score of 3.2933 further underscores our commitment to accuracy in content.

| Average Score | Brief Hospital Course | Discharge Instructions |
|---|---|---|
| Overall | 1.41 | 1.79 |
| Completeness | 2.48 | 3.01 |
| Correctness | 3.36 | 3.29 |
| Readability | 2.71 | - |

Table 6: Human Evaluation Results.

## 7 Conclusion

This paper presents a hybrid system developed by our team in participating in the "Discharge Me!" Challenge at the BioNLP 2024 Shared Task. We developed a hybrid system by combining extractive and abstractive summarization techniques. Our solution is triggered by the retrieval augmented generation (RAG) strategy that consists of a retriever to identify relevant information and a generator to generate the content. We fine-tuned GatorTron to recognize important problems, treatments, and lab tests from clinical notes as the retriever. We applied a clinical generative LLM, GatorTronGPT, as the generator to generate the target sections. Our approach was ranked 5th place among the participating teams, achieving an overall score of 0.284.

Using a fine-tuned encoder-only LLM, GatorTron, as the retriever, our system is able to capture important clinical concepts to reduce the input length and to focus the generator, GatorTronGPT, on those important clinical concepts. This strategy alleviated the challenge of token limitation in LLMs when dealing with clinical documents with extensive lengths. By integrating NER results from selected subset one with original text from the other selected subset, the input size was reduced by approximately 80% on average, which enabled the GatorTronGPT model to operate more efficiently and effectively.

The human evaluation results offer valuable insights into the quality of the generated discharge summaries. By conducting an average score from three clinicians, we got a more unbiased human-assessed performance of our generation pipeline. This process guides future enhancements to our model and data preprocessing methods. Overall, our study demonstrates the effectiveness of a hybrid approach that leverages both extractive and abstractive techniques in the generation of discharge summaries. The integration of NER and advanced generative modeling not only improves the manageability and performance of the task but also ensures the production of high-quality, contextually appropriate summaries.

## Limitations

We used the 2010 i2b2 dataset to fine-tune GatorTron to serve as the retriever. However, the challenge dataset was developed using clinical notes from a different source, which may hamper the performance of clinical concept extraction. The retriever only recognizes three types of concepts: problems, treatments, and lab tests. If GatorTron missed some key clinical concepts in the notes, GatorTronGPT may produce incomplete or inaccurate note sections. Future studies need to examine more advanced solutions.

## Acknowledgment

This study was partially supported by a Patient-Centered Outcomes Research Institute® (PCORI®) Award (ME-2018C3-14754), a grant from the National Cancer Institute, R01CA246418, grants from the National Institute on Aging, NIA R56AG069880, R01AG080624, R01AG083039, R01AG080991, R01AG084236, R01AG084178, R01AG076234, and R33AG062884, National Institute of Allergy and Infectious Diseases, NIAID R01AI172875, National Heart, Lung, and Blood Institute, NHLBI R01HL169277, the Cancer Informatics Shared Resource supported by the UF Health Cancer Center and the UF Clinical and Translational Science Institute Biomedical Informatics Program. The content is solely the responsibility of the authors and does not necessarily represent the official views of the funding institutions.

# A  Appendices

| Target Section | Selected Note Sections |
|---|---|
| Brief Hospital Course | **Subset 1:**<br>Physical Exam<br>Pertinent Results<br>Radiology report |
|  | **Subset 2:**<br>Chief Complaint<br>Major Surgical or Invasive Procedure<br>History of Present Illness<br>Diagnosis description<br>Discharge Disposition<br>Discharge Diagnosis<br>Discharge Condition |
| Discharge Instructions | **Subset1:**<br>Pertinent Results<br>Discharge Medications |
|  | **Subset2:**<br>Chief Complaint<br>Major Surgical or Invasive Procedure<br>History of Present Illness<br>Diagnosis description<br>Discharge Disposition<br>Discharge Diagnosis<br>Discharge Condition |

Table 7: Selected Note Sections for Target Sections

| Target Section | Example |
|---|---|
| Brief Hospital Course | <\|VIRTUAL_PROMPT\|> Input: **"Given the following concepts and text extracted from each section in a discharge summary, generate the section "Discharge Instructions". \n\nChief complaint: ... Discharge Instructions: \n** |
| Discharge Instructions. | <\|VIRTUAL_PROMPT\|> Input: **"Given the following concepts and text extracted from each section in a discharge summary, generate the section "Brief Hospital Course". \n\nChief complaint: ... Brief Hospital Course: \n** |

Table 8: Prompt Examples for Target Sections

| Target section | Ground Truth | GatorTronGPT Generated Text |
|---|---|---|
| Discharge Instructions | It was a pleasure caring for you at ___. You were admitted for a headache and for nausea. You were found to have too much fluid in your body (heart failure) and this was causing stress to your heart. We treated you with diuretics and with medications to improve the blood flow to your heart. Your blood pressure was also found to be too high so we added a new blood pressure medication. | Dear Ms. ___, It was a pleasure taking care of you at ___. You were admitted with nausea and vomiting and found to have a heart failure exacerbation. You were treated with diuretics and your symptoms improved. You also had high blood pressure. You were started on blood pressure medication" |
| Brief Hospital Course | Hospitalization Summary: Ms. ___ is an ___ yo ___ F with atrial fibrillation flutter on dabigatran, coronary artery disease, diastolic heart failure (LVEF 60% in ___, hypertension hyperlipidemia, and vascular dementia who presented with nausea and headache and was found to be in acute-on-chronic diastolic heart failure in the setting of NSTEMI. | Ms. ___ is an ___ with atrial fibrillation/flutter on dabigatran, coronary artery disease, diastolic heart failure (LVEF 60% in ___, hypertension/hyperlipidemia, and vascular dementia who presents with nausea and headache. # Nausea/vomiting: # Headache: # Hypertensive urgency: # NSTEMI: # Acute diastolic CHF exacerbation: # Hypertension: # Hyperlipidemia: |

Table 9: Comparison between GatorTronGPT Generated Summary and Ground Truth across Section